\documentclass[10pt,twocolumn,letterpaper]{article}

\usepackage[pagenumbers]{cvpr} 

\usepackage{graphicx}
\usepackage{amsmath}
\usepackage{amssymb}
\usepackage{booktabs}
\usepackage{array,multirow}
\usepackage{enumitem}

\usepackage[pagebackref,breaklinks,colorlinks]{hyperref}

\usepackage[capitalize]{cleveref}
\crefname{section}{Sec.}{Secs.}
\Crefname{section}{Section}{Sections}
\Crefname{table}{Table}{Tables}
\crefname{table}{Tab.}{Tabs.}

\def\cvprPaperID{17} 
\def\confName{CV4Animals workshop, CVPR}
\def\confYear{2023}

\begin{document}

\title{Towards Automatic Honey Bee Flower-Patch Assays \\with Paint Marking Re-Identification}

\newcommand{\hs}[0]{\hspace*{0.7cm}}

\author{\noindent\parbox{0.97\textwidth}{
\centering
Luke Meyers\textsuperscript{1,2}\hs
Rafael Meléndez-Ríos\textsuperscript{1}\hs
Josué Rodríguez Cordero\textsuperscript{1}\hs
Carlos Corrada Bravo\textsuperscript{1}\\ \vspace{0.5em}
Fanfan Noel\textsuperscript{1}\hs
José Agosto-Rivera\textsuperscript{1}\hs
Tugrul Giray\textsuperscript{1}\hs
Rémi Mégret\textsuperscript{1}
\\ \vspace{1.2em}
$^1$University of Puerto Rico, R\'io Piedras campus \qquad $^2$Seattle University}
}

\maketitle

\begin{abstract}
   In this paper, we show that paint markings are a feasible approach to automatize the analysis of behavioral assays involving honey bees in the field where marking has to be as lightweight as possible. We contribute a novel dataset for bees re-identification with paint-markings with 4392 images and 27 identities. Contrastive learning with a ResNet backbone and triplet loss led to identity representation features with almost perfect recognition in closed setting where identities are known in advance. Diverse experiments evaluate the capability to generalize to separate IDs, and show the impact of using different body parts for identification, such as using the unmarked abdomen only. In addition, we show the potential to fully automate the visit detection and provide preliminary results of compute time for future real-time deployment in the field on an edge device.
\end{abstract}

\section{Introduction}
\label{sec:intro}

Honey bees play an important role in our modern agriculture systems. As pollinators, the pollen they transfer between plants is critical to many crops' ability to produce fruits \cite{klein2007importance}. Developing methods to monitor and further study bees and other pollinators is important for maintaining and further understanding their contributions to our agricultural systems. Field observations can maximize pollination’s agricultural benefit and lead to recommendations of pollinator and crop management choices. In particular, in individualized foraging decisions and strategies assays, visiting bees can choose between artificial flowers of different morphologies and rewards \cite{fewell2000colony} \cite{giray2015effect}. Previous studies have shown variations of behavior at species level, which impact their choices when interacting with specific plants. However, to establish genetic linkage of behavioral traits demonstrated in the assay would require a much larger number of behaviors at individual level. 

Traditional experimental methods for such assays use visual inspection, which is limited to about 5 individual per day of experimentation due to humans limitations to monitor multiple individuals at once \cite{giray2015effect}. Computer Vision is a key element to improve the throughput of such studies, which in turn provide exciting challenges to push the state-of-the-art in insect monitoring:
(i) Open setup in the field, with poorly controlled lighting conditions; 
(ii) Need for accurate re-identification with no or lightweight markings to be performed during the experimentation itself;
(iii) Need for real-time analysis, as experimenters must physically catch individuals when specific conditions occur.

We propose several contributions towards these goals:
\vspace{-0.3ex}
\begin{itemize}[leftmargin=*,itemsep=0.2ex]
\item Methods and results showing the suitability of paintcode markings as a re-ID method, validated on a new contributed dataset of $4392$ high-quality images of honeybees comprising $27$ manually labeled IDs.
\item Method for detecting events during flower visits, validated on several videos taken in the field.   
\item Evaluation of the feasibility of integrating into an edge-computing system for the real-time acquisition and analysis in the field
\end{itemize}
The rest of the paper is organized as follows: after discussing related work (\cref{sec:prev}), we introduce the video collection system and visit detection approach (\cref{sec:pipeline}), present methods and results for re-ID (\cref{sec:reid}), and then provide preliminary results for a real-time implementation (\cref{sec:realtime}).

\section{Related Work}
\label{sec:prev}


Computer vision has shown strong promise in automatic detection and monitoring of animals and more specifically pollinators in a variety of environments. 

\cite{ratnayake2021tracking} have developed software that can effectively track insect movements in complex agricultural environments, using a combination of background subtraction and deep learning models to address challenges such as occlusion. Their Polytrack algorithm includes a feature to identify flower position and extract relevant pollination events including number of visits and duration. 

\cite{megret2019labelbee} developed a pipeline to characterize and identify bee behavior at the colony entrance based on pose estimation using deep convolutional networks \cite{rodriguez2018recognition}, and \cite{rodriguez2022automated} identified events
such as entering and leaving. The SLEAP \cite{pereira2022sleap} and DeepLabCut \cite{lauer_multi-animal_2022} platforms leverage such deep learning models for multi-animal pose estimation and tracking in videos for a large variety of species. 


In addition to short-term detection and tracking, recognizing the same individual across different timestamps or conditions, known as Re-Identification (reID), is an additional step to be able to monitor individual behavior over longer time periods. 

In the case of bees, automatic Re-ID has traditionally used tags or bar-codes. 
\cite{crall2015beetag} introduced the BEEtag system for individual bee tracking using fiducial markers, applied by \cite{smith2022long} with SLEAP pose detection \cite{pereira2022sleap} to track and identify individual bumble bees to study antennal activity. Similarly, \cite{boenisch2018tracking} proposed BeesBook, used by \cite{smith2022behavioral} to follow the lifetime behavior of thousands of bees.

Recent research has focused on markerless reID for various species, studied on datasets of seals \cite{nepovinnykh2022sealid}, tigers \cite{li2019atrw}, sea turtles \cite{papafitsoros2022seaturtleid}, and cattle \cite{bergamini2018multi}. 
In the case of bees, \cite{bozek2021markerless} proposed a method for tracking bees inside a colony, using similarity between bee instances to resolve crossings. \cite{borlinghaus2023purely} trained a FaceNet model for bumblebee reID trained on the triplet loss using images from \cite{tausch2020bumblebee}. Finally, \cite{chan2022honeybee} presented a hybrid dataset of honey bees containing a larger set of unmarked bees used for unsupervised pre-training, and a smaller set of tagged bees for fine-tuning with $126$ IDs for final evaluation. The effectiveness of the recognition from only the abdomen guided the design of our ablation study, comparing the effectiveness of full body images versus the abdomen and thorax crops (\cref{subsec:training_settings}).

To our knowledge no study has evaluated the possibility of automatic reID using paint markings, which has the potential of offering a more reliable performance than unmarked reID, while being faster and more lightweight than tagging. 

\section{Video collection in the field}
\label{sec:pipeline}

\subsection{Experimental setup}

The experimentation uses an artificial flower patch. Bees from a single hive are trained to forage at artificial flowers consisting of plexiglass squares with a small center well for nectar delivery. The assay considered in this work needs to evaluate the visits of each individual at each of the flowers for a specific number of visits, and then be captured to prepare them for a new experimental phase. This requires both detection of the visits, recognized when the bees place their head on top of the center well, as well as individual identification. Each bee needs to be marked on their first visit, which is a delicate operation that justifies the use of paint, which is much easier to apply than other markers (numbered tags or bar-codes) and can be decoded by both human observers and the machine.

\begin{figure}[t]
  \centering
   \includegraphics[width=\linewidth]{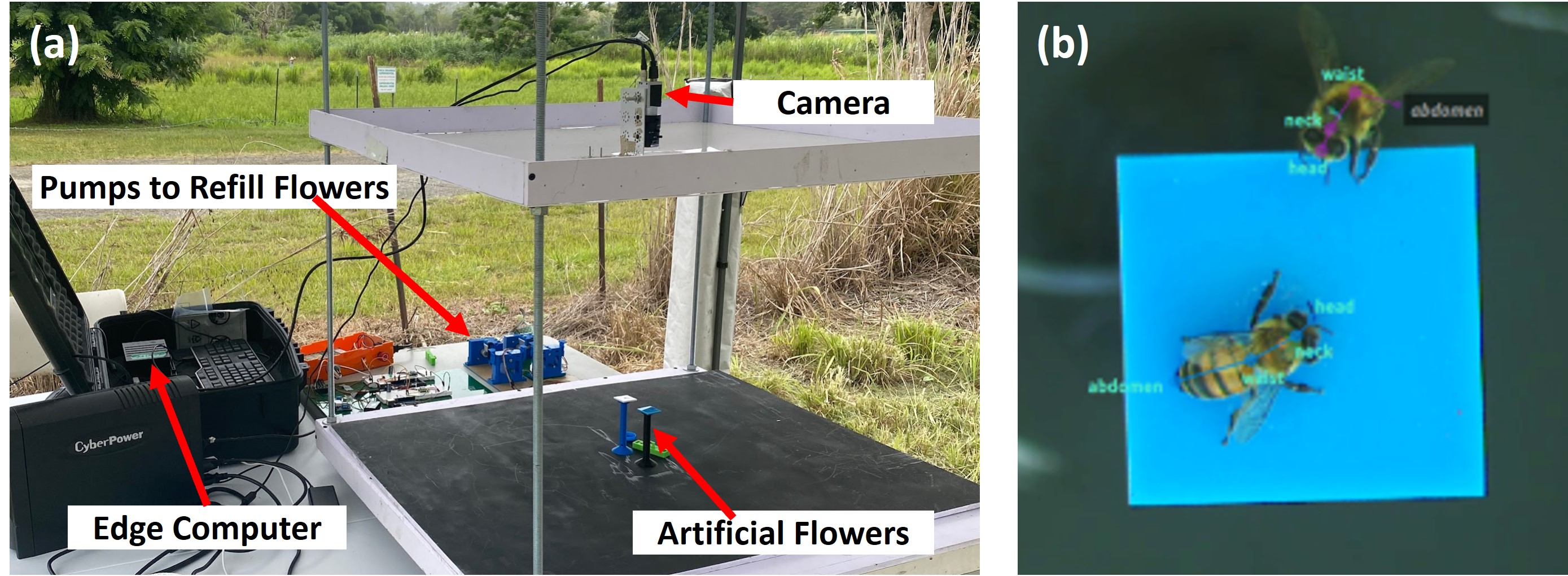}
\vspace{-0.6cm}
   \caption{(a) Overview of prototype flower-patch system. (b) Example of bee pose detection on one artificial flower.}
   \label{fig:overview}
\end{figure}

We used an experimental set up (Figure~\ref{fig:overview}) with two flowers, one blue and one white, with a Basler Gig-E camera mounted above the flower patch. White plexiglass was used to diffuse natural light and provided a surface to mount the camera using an L bracket next to a hole in the plexiglass. We tested the various frame rate and aperture settings of the camera to achieve optimum focus and lighting. The camera was
connected to a NVIDIA Jetson Xavier edge computing system to record using a GStreamer pipeline utilizing onboard hardware converters to increase speed. 
The video analysis pipeline discussed next was validated offline using pre-recorded videos, and is intended to be run real time in the future (see section~\ref{sec:realtime}).

\subsection{Visit detection}

Images were generated from videos collected over $4$ days of the experimental setup. During visits, researchers painted $1$ to $2$ small dots on each bee's thorax with various colors of enamel paint. 

SLEAP \cite{pereira2022sleap} pose detection software was used to annotate a frames across several videos, train a neural network for bee pose estimation. The skeletons used contain a head, neck, waist, and abdomen keypoints. The final model was generated using a total of $170$ annotated frames obtained incrementally by correcting the predictions from an initial model trained on $20$ frames. The model and tracking was then run on all videos to obtain short-term tracks.

Flower coordinates were generated by detecting their square shape using OpenCV. This provided the the position of each flower and their center well for the visit detection. 
A visit to drink nectar was detected when the head keypoint was within a certain distance from the center well in each frame. Temporally consecutive detections were agglomerated to produce drinking events with a start and end frame, associated to a flower ID and short-term track ID.

The evaluation of this module was performed on 3 videos, where a human annotated 98 drinking events. The automatic detection recalled 100\% of the events, but duplicated 9\% of them. 
(However, further review showed that in some cases bees had actually drank twice and moved slightly out of the well in between, suggesting the system can detect details that the human annotator had missed or ignored in the first visual inspection.)

\begin{figure}[t]
  \centering
   \includegraphics[width=1\linewidth]{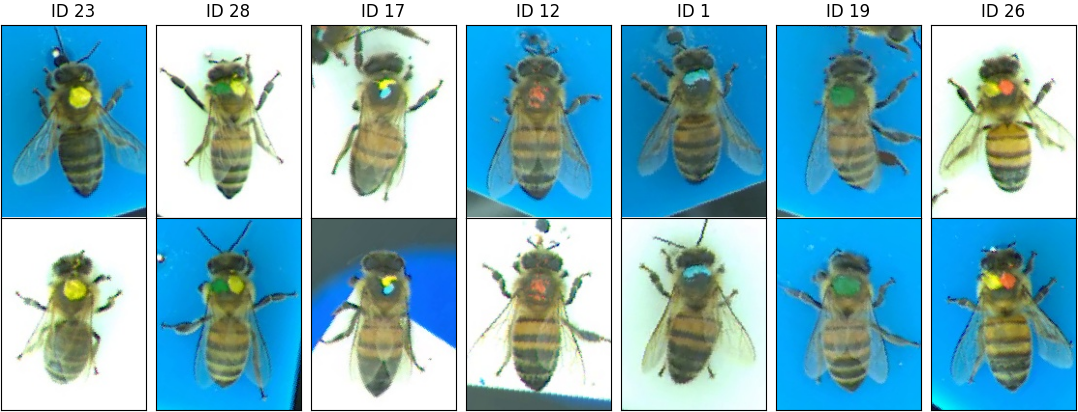}
\vspace{-0.6cm}
   \caption{Sample images from the flowerpatch Re-ID dataset. Two images are shown for each individual bee.}
   \label{fig:sample_img}
\end{figure}

\section{Bee ReID}
\label{sec:reid}

\subsection{Re-ID dataset}

From the tracks obtained, $150$ x $200$ pixel images of individual bees were extracted, centered on their waist keypoint and aligned based on the neck and waist angle. Photos with multiple bees were removed, as well as wrong orientation estimates and occluded bees.
Photos were then visually sorted based on individual ID and matching paint code. 
In creating the final dataset, tracks with less than $4$ images of a single ID, and IDs with less than $2$ tracks were removed.
The resulting flowerpatch Re-ID dataset contains $4392$ images with $27$ identities. It was obtained from $787$ short-term tracks, with an average of $5.6$ images sampled per track. Sample images are shown in Figure~\ref{fig:sample_img}. The dataset will be shared with the community.

\subsection{General approach}
\label{subsec:training_settings}

When deployed in practice, a re-ID model will not have access to data for all possible identities that it will encounter. To this end, we framed the problem as a representation learning problem, trained using a supervised contrastive learning approach \cite{khosla2020supervised}. Each input image is projected into a $128$ dimensional vector to be used as re-identification feature space using Euclidean distance. 

We considered two data splits: \textit{closed set setting} to evaluate the ideal supervised scenario; and \textit{open set setting} to measure how well the model can generalize to unseen identities.
(i) In the \textit{closed set setting}, all identities are present in both the train and test sets. We do so by first sorting all images by time and grouping them into tracks. The first $70$\% of tracks for each individual bee is assigned to the train set and the remaining to the test set. 
(ii) In the \textit{open set setting}, the train and test sets have distinct identities. We do so by assigning $60$\% of the IDs ($16$) to the train set and the remaining IDs ($11$) to the test set. For evaluation, we further split the test set into a \textit{reference} set and \textit{query} set by first sorting the test images by time and grouping them into tracks, and then assigning the first $25$\% to the reference set and the rest to the query set.

We furthermore distinguish four different versions of the input images: 1) \textit{aligned full body}; 2) \textit{aligned abdomen} (bottom crop); 3) \textit{aligned thorax} (top crop, which may include part of the head); and 4) \textit{unaligned full body}. The first three will help evaluate how information from different parts of the image affect model performance. The last one allows us to evaluate the usefulness of vertically aligning the bee images as an additional pre-processing step.

\subsection{Model Architecture and training}

We use a convolutional neural network (CNN) that embeds the input images into a $128$ dimensional feature space. This network consists of a ResNet$50$ backbone \cite{he2016deep} with bottleneck residual blocks, truncated after the \texttt{conv3} stage (after $22$ layers). The head is a fully connected layer with L$2$-normalization. 

The model was trained using the semi-hard triplet loss with a margin of $0.2$, optimized using Adam with a learning rate of $0.001$ for a maximum of $1000$ epochs and a patience of $100$. Dropout was applied during training both before and after the fully connected layer with rates of $0.5$ and $0.2$ respectively.

Unaligned full body uses $200$x$200$ pixel images trained with augmentation by a random rotation around the center.

\subsection{Evaluation}

Evaluation is performed using galleries and reporting the top-$1$ and top-$3$ Cumulative Matching Characteristic (CMC) metric. For each of the data splits, $10,000$ galleries are constructed as follows:

(i) For the \textit{closed set setting}, an anchor image image is randomly sampled from the test set, while a positive image along with $9$ negative images are randomly sampled from the train set.

(ii) For the \textit{open set setting}, an anchor image is randomly sampled form the query set, while a positive image along with $9$ negative images are randomly sampled from the reference set.

We additionally evaluate using $k$NN classifier trained on the complete reference data (train set for closed set setting, reference set for open set setting). 

A simple \textit{Pixel PCA} feature space is used as baseline: the raw image pixel RGB values flattened into a one-dimensional vector and projected into the principal components accounting for $95$\% of the variance from the train set.

\begin{table*}[ht!]
\centering
\begin{tabular}{|l |l  || l | l | l | l || l | l | l | l|} 
\hline               
& & \multicolumn{4}{|c||}{Closed Set Setting} & \multicolumn{4}{|c|}{Open Set Setting} \\
\cline{3-10}
 Features & Image Input & Top-1 & Top-3 & 1NN & 3NN & Top-1 & Top-3 & 1NN & 3NN   \\ 
\hline
\multirow{3}{*}{PixPCA} & Abdomen & 0.18 & 0.44 & 0.42 & 0.40 &  0.25 & 0.51 & 0.56 & 0.53 \\
\cline{2-10}
 &Thorax & 0.38 & 0.57 & 0.83 & 0.79 &  0.45 & 0.64 & 0.79 & 0.77 \\
\cline{2-10}
 &Full Img & 0.23 & 0.47 & 0.59  & 0.58 & 0.26 & 0.55 & 0.70 & 0.68 \\
\hline
\multirow{4}{*}{ResNet} &Abdomen &  0.94  &  0.990 & 0.90 & 0.91 &  0.74  &  0.93 & 0.94 & 0.94   \\ 
\cline{2-10}
 &Thorax &   0.992 &  \textbf{0.999} & 0.98 & \textbf{0.98}  & \textbf{0.90} &  \textbf{0.98} & 0.96 & \textbf{0.95} \\ 
\cline{2-10}
 &Full Img &   \textbf{0.993} &  \textbf{0.999} & \textbf{0.98} & \textbf{0.98}  &   0.87 & 0.92 &  \textbf{0.98} & 0.92   \\ 
\cline{2-10}
 &Full Img Unaligned & 0.98  & 0.997 & 0.95 & 0.95  & 0.78  & 0.95 & 0.93 & 0.91 \\ 
\hline
\end{tabular}
\caption{CMC Scores and kNN Accuracies for re-identification in closed and open set settings.}
\label{table:cmc_scores}
\end{table*}

\subsection{Results}
\label{sec:results}

\paragraph{Closed Set Setting}

\cref{table:cmc_scores} shows that in the closed set setting, the baseline which uses pixel data without any model training can reach an accuracy of almost $0.83$ $1$-NN accuracy when using the thorax, but only $0.38$ Top-$1$ accuracy, which indicates a large intra-class variability that could be compensated with enough training data.
The contrastively trained ResNet model performs very well regardless of input type, with the thorax crop or full image reaching almost perfect performance. The abdomen performance is lower, but still demonstrates the potential of using the unmarked abdomen for Re-ID, which aligns with previous work of \cite{chan2022honeybee}.

Results for ResNet trained on un-aligned images were only slightly lower than for aligned images. Thus body alignment did not lead to a large improvement in the closed set setting: augmentation during training seemed to account for the variability in the input.

\paragraph{Open Set Setting}

\cref{table:cmc_scores} shows that in the open set setting, we see a larger gap between abdomen crop and thorax crop results, showing the importance of paint marking to obtain reliable re-identification in that context. It would be interesting for future work to evaluate the impact of using larger training sets on this gap. We note that all individual paint colors in the test set IDs are also present in the train set IDs, although in slightly different shapes and position, which may account for why thorax performed well.

Performance for the model trained on un-aligned images was lower than for aligned images. Thus, unlike in the closed set setting, vertically aligning the images was needed in the open set to help generalize to new IDs.

\section{Towards real-time}
\label{sec:realtime}

Real-time analysis is critical for taking actionable decisions in the field during behavioral assays. In order to move from offline processing to real-time processing in the field, where the system may need to be battery powered, the hardware constraints of the system such as power consumption and computing capabilities impose additional limitations. Using the same NVIDIA Jetson Xavier AGX platform as was used for data collection, we trained a Yolov$5$ model \cite{yolov5} to detect the bees in our videos and benchmarked its performance on the videos cropped to $1184$ x $1184$ pixels. 

As a baseline, frame-by-frame PyTorch inference took $60$ms per frame ($17$ fps). Converting the model to a TensorRT engine and running inference within an NVIDIA DeepStream pipeline resulted in $13$ms per frame (59 fps). Adding a tracker (NvDCF) and an Apache Kafka \cite{apache_kafka} message broker to export the trajectories added $1$ms per frame. These preliminary results demonstrate that although a naive implementation of the models may not reach the $20$ fps target, it is feasible to significantly exceed this framerate for bee detection and tracking with the proper computational optimizations, leaving capacity to add the head pose estimation, re-id modules and visit detection.

\section{Conclusion}
\label{sec:conclusion}

In this paper, we presented a comprehensive set of methods to tackle the problem of individual behavior monitoring in flower patch assays in the field.
We presented a novel dataset of honey bees with paint marks and trained a deep learning model for reID. Our results show that paint marks is a viable light-weight alternative to more traditional tagging, thus facilitating the automatic re-identification of individuals during field experiments. These results showed a strong benefit of vertically aligning images in the open set setting, but was not as strong in the closed set. They also showed promising performance of using only the unmarked abdomen, suggesting the prospect of usable unmarked Re-ID in the future with improved models. The CNN based detection was of good quality, leading to accurate drinking event detections. Integration of the various modules into a real-time system is in progress, with the potential to enable high-throughput individual honey bee behavior characterization in the field.


\section*{Acknowledgement}
\label{sec:acknowledgment}

This work is supported by grant no. 2021-67014-34999 from the USDA National Institute of Food and Agriculture. This material is based upon work supported by the National Science Foundation PIRE Award 1545803 and NSF IQ-BIO-REU Award 1852259. T. G. acknowledges NSF-HRD award 1736019 that provided funds for the purchase of bees. This work used the High-Performance Computing facility (HPC) Award Number 5P20GM103475 supported by an Institutional Development Award (IDeA) from the National Institute of General Medical Sciences of the National Institutes of Health under grant number P20GM103475.

{\small
\bibliographystyle{ieee_fullname}
\bibliography{PaperFinal}
}

\end{document}